\documentclass[sigconf]{acmart}

\usepackage{amsmath,amsfonts,bm}

\def\eqref#1{equation~\ref{#1}}

\def\1{\bm{1}}

\DeclareMathAlphabet{\mathsfit}{\encodingdefault}{\sfdefault}{m}{sl}
\SetMathAlphabet{\mathsfit}{bold}{\encodingdefault}{\sfdefault}{bx}{n}

\usepackage{hyperref}
\usepackage{url}
\usepackage{graphicx}
\usepackage{amsmath,amsthm}
\usepackage{booktabs}
\usepackage{thmtools,thm-restate}
\usepackage{mathtools}
\usepackage{subfig}
\usepackage{multirow}
\usepackage{colortbl}
\usepackage{enumitem}
\usepackage{color}
\usepackage{makecell}
\usepackage[linesnumbered,boxed,ruled,vlined]{algorithm2e}
\usepackage{listings}

\usepackage{flushend}
\usepackage{makecell}
\usepackage[normalem]{ulem}
\usepackage{CJKutf8}
\usepackage[utf8]{inputenc}
\usepackage{balance}

\newcommand{\hide}[1]{}

\newcommand{\vpara}[1]{\vspace{0.2cm}\noindent\textbf{#1 }}

\usepackage{flushend}
\usepackage[normalem]{ulem}

\usepackage{xspace}

\definecolor{lightgreen}{RGB}{197,224,180}
\definecolor{lightblue}{RGB}{222,235,247}
\definecolor{lightpurple}{RGB}{238,229,241}
\definecolor{lightorg}{RGB}{251,229,214}

\AtBeginDocument{%
  \providecommand\BibTeX{{%
    \normalfont B\kern-0.5em{\scshape i\kern-0.25em b}\kern-0.8em\TeX}}}

\setcopyright{acmcopyright}
\copyrightyear{2021}
\acmYear{2021}

\begin{document}

\fancyhead{}

\title{M6: A Chinese Multimodal Pretrainer}

\author[J. Lin*, R. Men*, A. Yang*, C.Zhou, Y. Zhang, P. Wang, J. Zhou, J.Tang, H. Yang]{
    Junyang Lin$^{1*}$, Rui Men$^{1*}$, An Yang$^{1*}$, Chang Zhou$^1$, Ming Ding$^2$, Yichang Zhang$^1$, Peng Wang$^1$, Ang Wang$^1$, Le Jiang$^1$, Xianyan Jia$^1$, Jie Zhang$^1$, Jianwei Zhang$^1$, Xu Zou$^2$, Zhikang Li$^1$, Xiaodong Deng$^1$, Jie Liu$^1$, Jinbao Xue$^1$, Huiling Zhou$^1$, Jianxin Ma$^1$, Jin Yu$^1$, Yong Li$^1$, Wei Lin$^1$, \\ Jingren Zhou$^1$, Jie Tang$^{2\dagger}$, Hongxia Yang$^{1\dagger}$
}
\affiliation{
    $^1$Alibaba Group\country{China} \\
    $^2$Tsinghua University\country{China}
}
\email{
    {junyang.ljy, menrui.mr, ya235025, ericzhou.zc, yichang.zyc, zheluo.wp}@alibaba-inc.com
}
\email{
    {wangang.wa, jiangle.jl, xianyan.xianyanjia, wanglin.zj, zhangjianwei.zjw}@alibaba-inc.com
}
\email{
    {zhikang.lzk, xiaodongdeng.dxd, sanshuai.lj, zhiji.xjb, zhule.zhl, jason.mjx, kola.yu}@alibaba-inc.com
}
\email{
    {jiufeng.ly, weilin.lw, jingren.zhou, yang.yhx}@alibaba-inc.com
}
\email{
    {dm18, zoux18}@mails.tsinghua.edu.cn,jietang@tsinghua.edu.cn
}



\begin{abstract}

In this work, we construct the largest dataset for multimodal pretraining in Chinese, which consists of over 1.9TB images and 292GB texts that cover a wide range of domains. 
We propose a cross-modal pretraining method called \textbf{M6}, referring to \textbf{M}ulti-\textbf{M}odality to \textbf{M}ulti-\textbf{M}odality \textbf{M}ultitask \textbf{M}ega-transformer, for unified pretraining on the data of single modality and multiple modalities. 
We scale the model size up to 10 billion and \textbf{100 billion} parameters, and build the largest pretrained model in Chinese. 
We apply the model to a series of downstream applications, and demonstrate its outstanding performance in comparison with strong baselines. 
Furthermore, we specifically design a downstream task of text-guided image generation, and show that the finetuned M6 can create high-quality images with high resolution and abundant details.

\end{abstract}

\keywords{Multimodal Pretraining; Multitask; Text-to-Image Generation}


\maketitle

\section{Introduction}
\label{sec:intro}

Pretraining has become a focus in the research in natural language processing (NLP)~\citep{gpt2, gpt3, xlnet, albert, roberta, convbert, mass, structbert, unilm, unilmv2, deberta}. 
The recent GPT-3 with over 175 billion parameters demonstrates that large models trained on big data have extremely large capacity and it can outperform the state-of-the-arts in downstream tasks especially in the zero-shot setting. 
Also, the rapid development of pretraining in NLP sparkles cross-modal pretraining. 
A number of studies~\citep{vilbert, vl-bert, vlp, unicoder-vl, uniter, oscar, villa, vilbert-mt, interbert, pixelbert} have created new state-of-the-art performances for various cross-modal downstream tasks. 

A pity is that most recent studies focus on the pretraining on English data. 
There are lack of both large-scale datasets in Chinese and large-scale models pretrained on the data of Chinese. 
Therefore, in this work, we develop a large-scale dataset M6-Corpus, which consists of over 1.9TB images and 292GB texts. 
To the best of our knowledge, this is the largest dataset in Chinese for pretraining in both multimodality and natural language. 
The dataset collected from the webpages consists of different types of data and covers a large scale of domains, including encyclopedia, question answering, forum discussion, product description, etc. 
Also, we design sophisticated cleaning procedures to ensure that the data are of high quality. 

Furthermore, in order to sufficiently leverage such a large amount of high-quality data, we propose to build an extremely large model that can process data of multiple modalities and adapt to different types of downstream tasks. 
Thus we propose a novel model called M6, referring to MultiModality-to-MultiModality Multitask Mega-transformer. 
The model is based on the transformer, and it is pretrained with multiple tasks. 
Pretraining endows the model with the capability of single-modality and multimodality understanding and generation. 
Based on the architecture of M6, we build \textit{M6-10B} and \textit{M6-100B}, which are scaled up to 10 billion and 100 billion parameters respectively. 
To be more specific, \textit{M6-100B} is the recent largest model pretrained on Chinese data. 
We apply the model to a series of downstream applications, including product description generation, visual question answering, community question answering, Chinese poem generation, etc., and our experimental results show that M6 outperforms a series of strong baselines. 

Another contribution of this work is that we first incorporate pretraining with text-to-image generation. 
Following \citet{dalle}, we leverage a two-stage framework for image generation. 
To be more specific, we apply a trained vector-quantized generative adversarial network to representing images with discrete image codes, and we then use the pretrained M6 to learn the relations between texts and codes. Such learning can bridge the two modalities and enables controllable text-to-image generation. 

\renewcommand{\thefootnote}{\fnsymbol{footnote}}
\footnotetext[1]{Equal contribution.}
\footnotetext[2]{Corresponding author.}
\renewcommand{\thefootnote}{\arabic{footnote}}

To summarize, the contributions of M6 are as follows:
\begin{itemize}
    \item We collect and build the largest Chinese multi-modal pre-training data in  industry, which includes 300GB texts and 2TB images. 
    \item We propose M6 for multimodal pretraining in Chinese, and we scale the model size to up to 10 and 100 billion parameters. Both M6-10B and M6-100B are the recent largest multimodal pretrained model. 
    \item M6 is versatile and exceeds strong baselines by 11.8\% in VQA, 18.4 in image captioning, and 10.3\% in image-text matching. Furthermore M6 is able to generate high-quality images. 
    \item With carefully designed large-scale distributed training optimizations, M6 has obvious advantages in training speed and greatly reduces training costs, creating the possibility for more widespread use of multi-modal pretraining.
\end{itemize}



\section{Dataset}

\begin{table*}[t]
  \caption{Statistics of our pretraining dataset. We demonstrate the sources of our data, and we calculate the number of images, tokens, and passages, the average length, as well as the size of images and texts.} 
  \label{tab:pretrain_data}
  \begin{tabular}{cccccccc}
    \toprule
        Source   & Modality & Images (M) & Tokens (B) & Passages (M) & Avg. Length & Image Size (TB) & Text Size (GB) \\
        \midrule
        Encyclopedia  & Plain-text & - & 31.4 & 34.0 & 923.5 & - & 65.1   \\
        Community QA  & Plain-text & - & 13.9 & 113.0 & 123.0 & - & 28.8   \\
        Forum discussion  & Plain-text & - & 8.7 & 39.0 & 223.1 & - & 18.0  \\
        Common Crawl  & Plain-text & - & 40.3 & 108.7 & 370.7 & - & 83.3   \\
        \midrule
        Encyclopedia  & Image \& Text & 6.5 & 7.9 & 10.4 & 759.6 & 0.1 & 15.0 \\
        Crawled Webpages  & Image \& Text & 46.0 & 9.1 & 106.0 & 85.8 & 1.5 & 70.0  \\
        E-commerce  & Image \& Text & 8.0 & 0.5 & 8.5 & 62.1 & 0.3 & 12.2 \\ 
        \midrule
        Total   & - & 60.5 & 111.8 & 419.6 & 266.4 & 1.9 & 292.4 \\
    \bottomrule
  \end{tabular}
\end{table*}

\begin{table}[t]
  \caption{Comparison with the existing large-scale Chinese corpora for pretraining. Our dataset is the largest dataset for Chinese pretraining. The size of texts is larger than that of the existing datasets, and the size of images is even larger than that of ImageNet. } 
  \label{tab:dataset_comp}
  \begin{tabular}{cccc}
    \toprule
        Dataset   & Text Size (GB) & Image Size (GB) & Multidomain   \\
        \midrule
        CN-Wikipedia & 1.6 & $\times$ & $\times$ \\
        THUCTC   & 2.2 & $\times$ & $\times$ \\
        HFL & 21.6 & $\times$ & \checkmark \\
        CLUE Corpus   & 100.0 & $\times$ & \checkmark \\
        ImageNet & $\times$ & $\sim$1000 & \checkmark \\
        M6-Corpus    & 292.4 & 1900 & \checkmark \\
    \bottomrule
  \end{tabular}
\end{table}

\begin{figure*}
    \centering
    \includegraphics[width=0.95\textwidth]{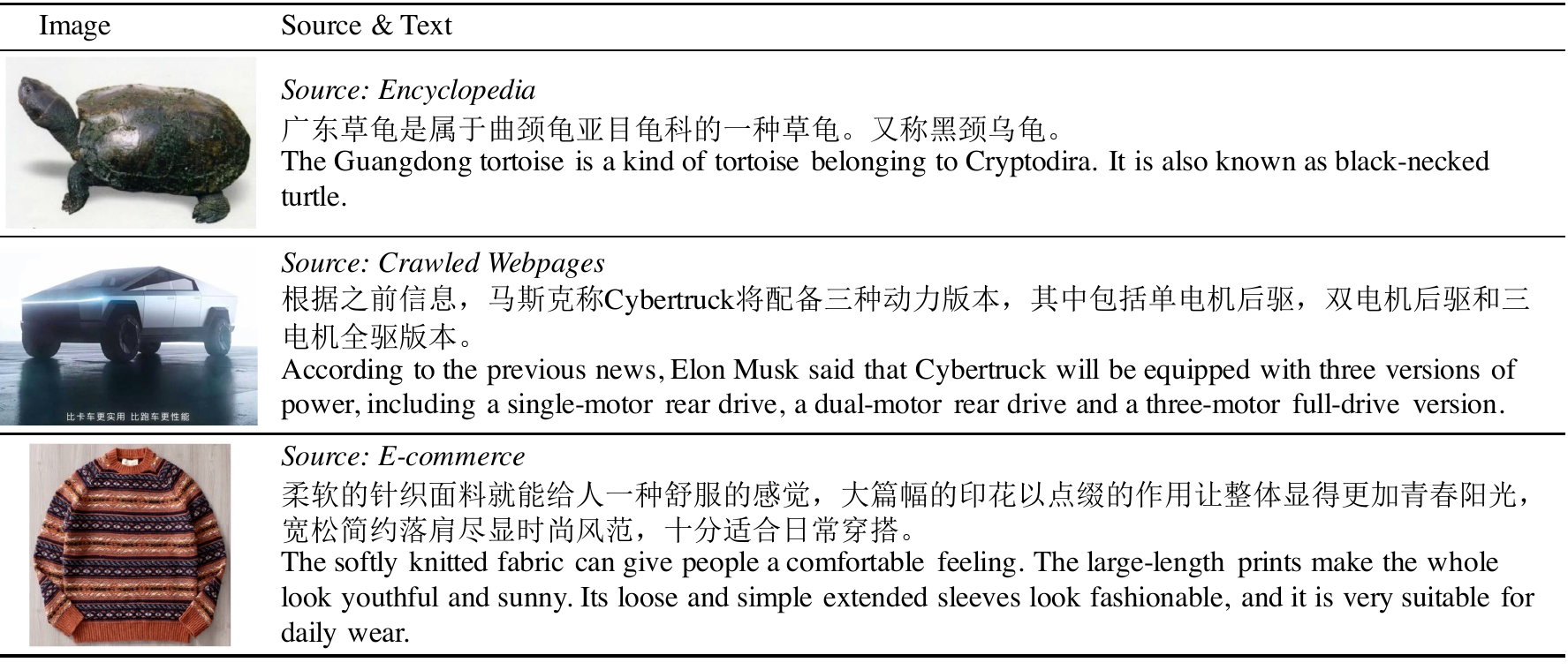}
    \caption{Examples of the multimodal data of M6-Corpus. We demonstrate three cases that belong to different categories, including encyclopedia, crawled webpages, and product description.}
    \label{fig:image_case}
\end{figure*}

\begin{figure*}
    \centering
    \includegraphics[width=0.95\textwidth]{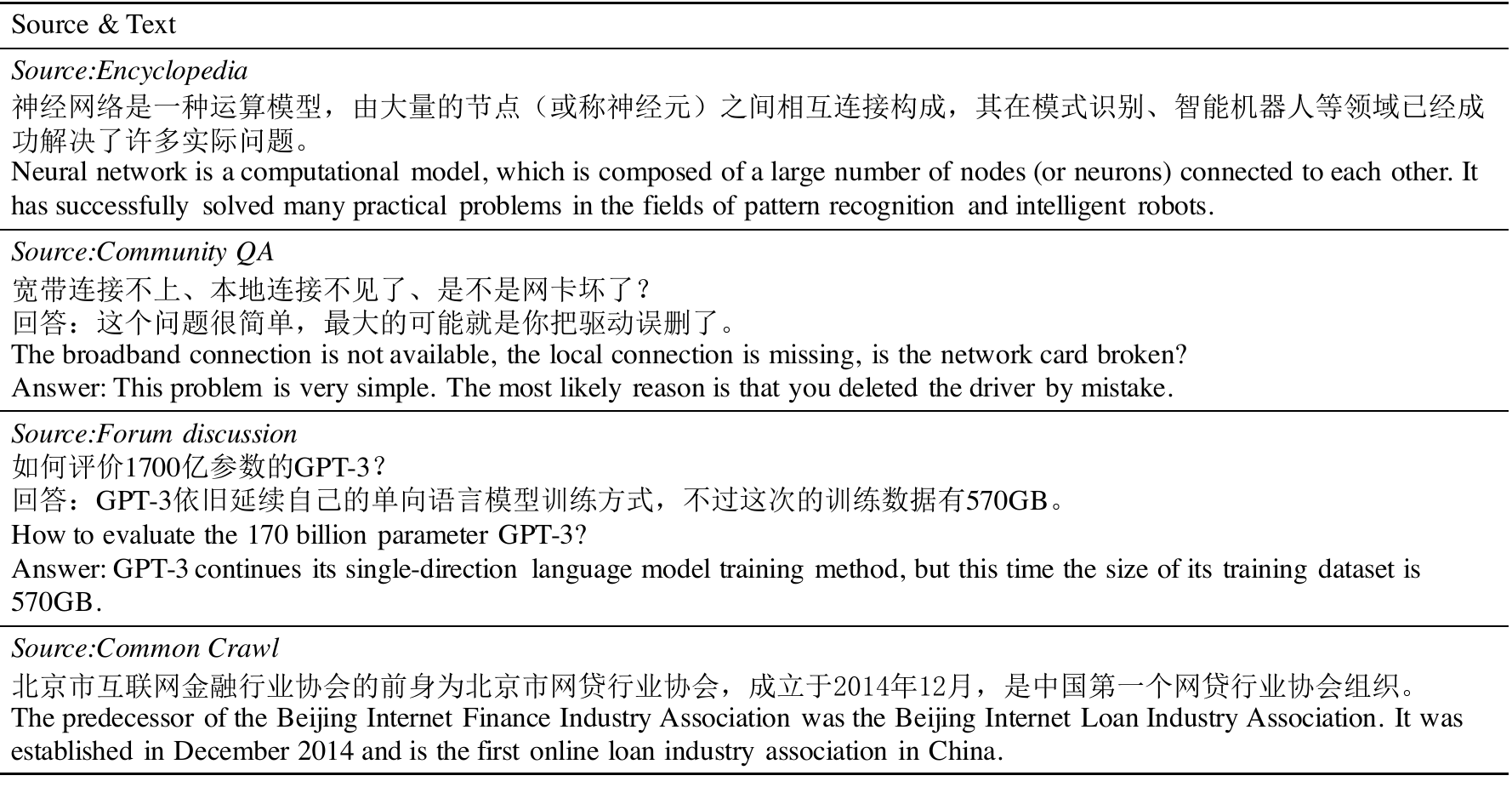}
    \caption{Examples of the plain text data of M6-Corpus. We demonstrate three cases that belong to different categories, including encyclopedia, community QA, forum discussion, and common crawl.}
    \label{fig:text_case}
\end{figure*}

\begin{table*}[t]
  \caption{Statistics of the pretraining dataset. We demonstrate the sources of our data, and we calculate the number of images, tokens, and passages, as well as the size of images and texts.} 
  \label{tab:pretrain_data}
  \begin{tabular}{cccccc}
    \toprule
        Source  (M) & Images(M) & Tokens (B) & Passages (M) & 
        Image Size (TB) & Text Size (GB) \\
        \midrule
        Encyclopedia    & 6.5 & 7.9 & 10.4 & 0.1 & 15.0 \\
        Webpages & 46.0 & 9.1 & 106.0 & 1.5 & 70.0  \\
        E-commerce  & 8.0 & 0.5 & 8.5 & 0.3 & 12.2 \\ 
        \midrule
        Total & 60.5 & 17.5 & 124.9 & 1.9 & 97.2 \\
    \bottomrule
  \end{tabular}
\end{table*}

We collect and develop the largest multi-modality and text dataset in Chinese for now, which is one of the key contributions of this paper.
In this section, we first identify the limitations of existing datasets and then describe the construction and preprocessing procedure of our proposed dataset.



\subsection{Existing Datasets}
The construction of large-scale corpus with high quality and domain coverage is crucial to Chinese pretraining. 
In early previous works, the Chinese Wikipedia\footnote{https://dumps.wikimedia.org/zhwiki/latest/} is one of the most frequently used datasets to train Chinese language models. 
It contains 1.6GB texts (around 0.4B tokens) covering around 1M encyclopedia entries.
Another corpus with a comparable size is the THUCTC\cite{thuctc} dataset, which includes 740K news articles. 
However, with the rapidly increasing capacity of recent language models, the scale of these existing datasets is clearly insufficient.
Recently, \citet{hfl_bert} employs unreleased extended data that are 10 times larger than the CN-Wikipedia to pretrain their Chinese language model. \citet{cluecorpus2020} released a 100GB corpus named CLUECorpus2020, which is retrieved from the multilingual Common Crawl dataset.
However, the scale of the datasets is still insufficient to facilitate super large-scale pretraining compared with existing English pretrained models.
For example, GPT-3 contains 175B parameters and is trained on 570GB texts.
Meanwhile, the dataset should contain image-text pairs rather than plain texts for multi-modal pretraining.

\subsection{Standards for a High-quality Dataset}
To perform large-scale multi-modal pretraining and learn complex world knowledge in Chinese, the dataset is highly required to provide both plain texts and image-text pairs on super large scale, covering a wide range of domains.
In order to perform large-scale multi-modal pretraining in Chinese, we focus on the construction of large-scale datasets in Chinese. 
Specifically, while we unify our pretraining for both natural language and multimodalities, we construct large datasets of both plain texts and image-text pairs. 
We are interested in obtaining large-scale data that covers a wide range of domains, so that it is possible for the model to learn the complex world knowledge of different fields. 
Also, we aim to collect data of multiple modalities for the cross-modal pretraining.
This raises the difficulty for the construction of a large-scale dataset as the data for multimodal pretraining are usually image-text pairs, where in each pair the text provides a detailed description of a fraction of the image. 

Though there are a tremendous amount of text resources and images on the world wide web, the corpus for multimodal pretraining is assumed to be better when satisfying the following properties:
(1). the sentences should be fluent natural language within a normal length, and should not contain meaningless tokens, such as markups, duplicate punctuation marks, random combinations of characters, etc.;  
(2). the images should be natural and realistic, and the resolutions of the images need to be identifiable by humans;
(3). both the texts and images should not contain illegal content, such as pornography, violence, etc.;
(4). the images and texts should be semantically relevant;
(5). the datasets should cover a wide range of fields, say sports, politics, science, etc., and therefore it can endow the model with sufficient world knowledge. 

\subsection{Dataset Construction}
Based on the requirements above, we collect data of both plain texts and image-text pairs.
There are different types of data, including encyclopedia, crawled webpage, community question answering, forum, product description, etc. 
We present the details in Table~\ref{tab:pretrain_data}. 
The collected corpus consists of both plain-texts and image-text pairs, which is compatible with the designed text-only and multimodal pretraining tasks.
Also, the data has a large coverage over domains, such as science, entertainment, sports, politics, commonsense of life, etc. 
We have also compared some characteristics of our corpus with existing datasets used for Chinese pretraining in Table~\ref{tab:dataset_comp}.
The size of our dataset is much larger than the previous ones.
To our knowledge, this is the first large-scale, multimodal and multidomain corpus for Chinese pretraining.

We implement sophisticated preprocessing to obtain clean data.  
For text data, we first remove HTML markups and duplicate punctuation marks, and we only reserve characters and punctuation marks that are in Chinese and English.
We remove the topics that are shorter than 5 characters and contents shorter than 15 characters.
We further apply  in-house spam detection to remove sentences that contain words related to certain political issues, pornography, or words in the list of dirty, naughty, and other bad words. 
In order to preserve the linguistic acceptance of the texts, we implement a language model to evaluate their perplexities, and sentences with high perplexities are discarded.
Only images with at least 5000 pixels are reserved for pretraining.
A sequence of classifiers and heuristic rules are applied to filter out images containing illegal content.
We also use a pretrained image scorer to evaluate the qualities of images.
For images and texts in crawled webpages, we only consider images and  their surrounding text as relevant image-text pairs.
Other sentences in the webpages are discarded.

\section{M6 Framework}

 \begin{figure*}[tb]
    \centering
    \includegraphics[width=1.0\linewidth]{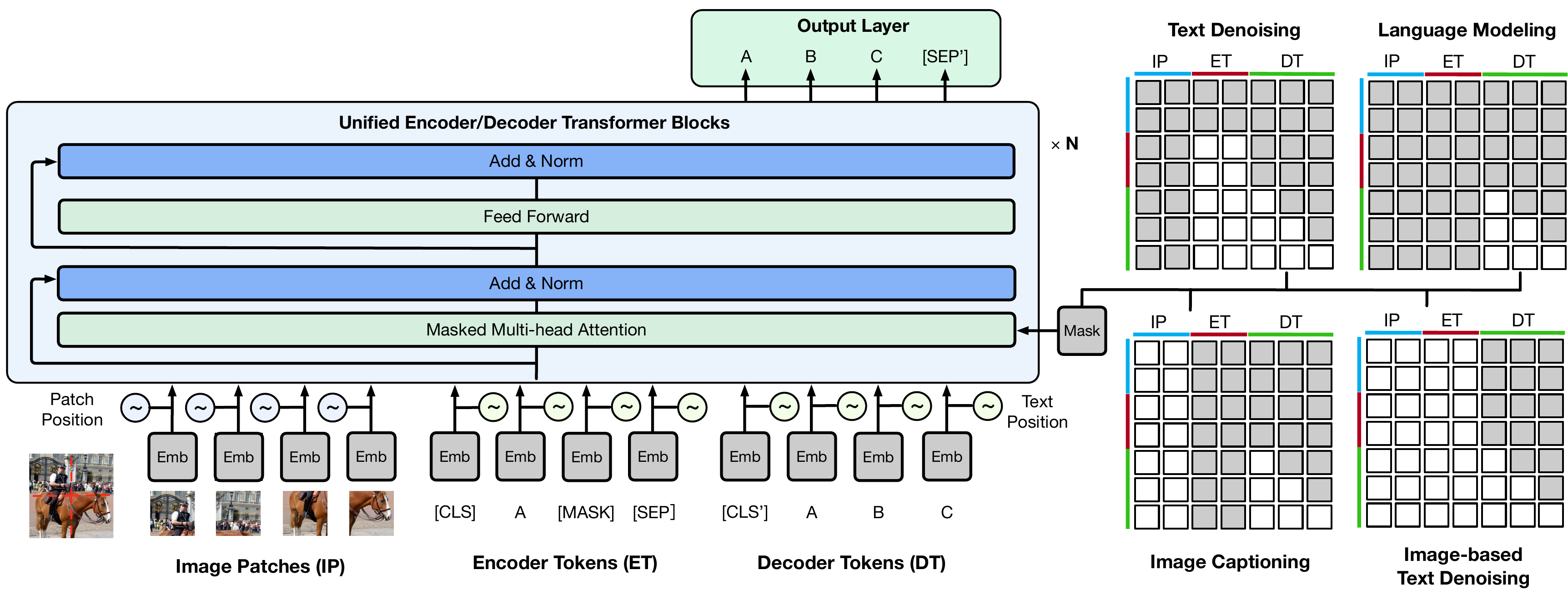}
    \caption{An overview of the pretraining tasks for M6. The design of masking strategies allows the learning of different tasks under the same framework. M6 is pretrained with image-based text denoising, image captioning, text denoising, and language modeling.}
    \label{fig:pretrain_tasks}
  \end{figure*}

Multimodal pretraining leverages both the power of self-attention-based transformer architecture and pretraining on large-scale data. 
We endeavor to endow the model with strong capability of cross-modal understanding and generation. 
In this section, we describe the details of our proposed pretrained model \textbf{M6}, which refers to \textbf{M}ulti-\textbf{M}odality-to-\textbf{M}ulti-\textbf{M}odality \textbf{M}ultitask \textbf{M}ega-transformer.

\subsection{Visual and Linguistic Inputs}\label{image_features}
The mainstream multimodal pretraining methods transform images to feature sequences via object detection. 
However, the performance of the object detectors as well as the expressivity of their backbones strongly impact the final performance of the pretrained models in the downstream tasks. 
We observe that a large proportion of the images contain only a few objects. 
Take the images of the data of e-commerce as an example. 
We randomly sample 1M images and perform object detection on the images. 
The results show that over 90\% of the images contain fewer than 5 objects. 
Also, the objects have high overlapping with each other. 
To alleviate such influence, we turn to a simple but effective solution following \citet{fashionbert} and \citet{vit}. 
In general, we split an image into patches and extract features of the 2D patches with a trained feature extractor, say ResNet-50. 
Then we line up the representations to a sequence by their positions.


The processing of the input word sequence is much simpler. We follow the similar preprocessing procedures in the previous work~\citep{uniter, oscar, villa}. 
We apply WordPiece~\citep{gnmt, bpe} and masking to the word sequence and embed them with an embedding layer, following BERT~\citep{bert}. 

\subsection{Unified Encoder-Decoder}
We integrate the image embeddings $e^i$ and the word embeddings $e^t$ into the cross-modal embedding sequence $e = \{e^i, e^t\}$. We send the sequence to the transformer backbone for high-level feature extraction. 
To differ their representations, we add corresponding segment embeddings for different modalities. 
Specifically, we leverage the self-attention-based transformer blocks for our unified cross-modal representation learning. To be more specific, the building block is identical to that of BERT or GPT, which consists of self attention and point-wise feed-forward network (FFN).
On top of the transformer backbone, we add an output layer for word prediction, and thus we tie its weights to those of the embedding layer. 

In the unified framework, we use different masking strategies to enable encoding and decoding. 
The input is segmented into three parts, including visual inputs, masked linguistic inputs, and complete linguistic inputs. 
We apply bidirectional masking to both the visual inputs and masked linguistic inputs, and we apply causal masking to the complete linguistic inputs. 
Thus the model is allowed to encode and decode in the same framework.

\subsection{Pretraining Methods}

We pretrain the model with the multitask setup, including text-to-text transfer, image-to-text transfer, and multimodality-to-text transfer. Thus the model can process information of different modalities and perform both single-modal and cross-modal understanding and generation. 


\vpara{Text-to-text Transfer}
As demonstrated in Figure~\ref{fig:pretrain_tasks}, the model learns to perform text denoising and language modeling in the setting of text-to-text transfer. 
In text denoising, we mask the input text by a proportion, which is 15\% in practice following BERT~\citep{bert}. 
Specifically, we mask a continuous span of text with a single mask, and the model should learn to decode the whole sequence. This encourages the model to learn both recovering and length predicting. 
Besides, in order to improve the model ability in generation, we add a setup of language modeling, where the encoder receives no inputs and the decoder learns to generate words based on the previous context. 

\vpara{Image-to-text transfer}
Image-to-text transfer is similar to image captioning, where the model receives the visual information as the input, and learns to generate a corresponding description. 
In this setting, we add the aforementioned patch feature sequence to the input and leave the masked input blank. The model encodes the patch features, and decodes the corresponding text. 

\vpara{Multimodality-to-text transfer}
Based on the setup of image-to-text transfer, we additionally add masked linguistic inputs, and thus the model should learn to generate the target text based on both the visual information and the noised linguistic information. 
This task allows the model to adapt to the downstream tasks with both visual and linguistic inputs.

\begin{table}
  \caption{Model sizes of M6. $n_{layers}$ is the number of transformer layers. $d_{model}$ is the dimension of hidden states in each layer. $n_{heads}$ is the number of attention heads in each layer. $n_{experts}$ is the number of experts. The M6-100B model employs multiple experts to scale up parameters to 100 billion. $n_{param}$ is the number of all parameters.} 
  \label{tab:model size}
  \begin{tabular}{ccccccc}
    \toprule
        Models & $n_{layers}$ & $d_{model}$ &  $n_{heads}$  & $n_{experts}$ & $n_{param}$    \\
        \midrule
           M6-base & 24 & 1024 & 16 & 1 & 327M \\
           M6-10B & 50 & 4096 & 128 & 1 & 10B \\
           M6-100B & 24 & 1024 & 16 & 1024 & 100B \\
    \bottomrule
  \end{tabular}
\end{table}

\subsection{Scaling up to 10 and 100 Billion Parameters}
We scale up the model size to 10 billion parameters and 100 billion parameters, which are named M6-10B and M6-100B. 
The increase in model size provides a much larger capacity for the model that it can learn knowledge from more data. 
For the construction of M6-10B, we simply scale up the model by hyperparameter tuning. 
To be more specific, we increase the size of hidden states and the number of layers. 
To better leverage GPU memory, we apply mixed-precision training and activation checkpointing to save memory. 
Still, the model cannot be fit into one single GPU, and thus we use model parallelism to split the feed-forward networks and attention heads to multiple GPUs following the implementation of Megatron-LM~\citep{megatron}. 

However, directly scaling up to M6-100B is much more difficult as there are more challenges for the computation resources. 
Alternatively, inspired by the recent progress in sparse activations~\citep{moe, gshard, switch}, we combine Mixture-of-Experts (MoE) with M6 to build the version of 100 billion parameters. 
Note that the original MoE requires mesh-tensorflow as well as TPUs. 
This sets limits for a number of researchers without such resources. 
Thus we implement the M6-100B with MoE with our in-house framework Whale~\citep{whale} to perform model parallelism with GPUs. 
We demonstrate the key statistics of the models of different scales in Table~\ref{tab:model size}. 

Specifically, different from the conventional FFN layer, the MoE layer is a parallel combination of multiple FFN layers, each of which acts as an expert. This is also called expert parallelism. 
The model first learns a sparse gating network to route the tokens to specific experts. Thus each token is only sent to a small set of experts and the computation can be much less compared with that in dense models. This kind of model is highly efficient as it realizes data parallelism and expert parallelism across workers.  The computation of MoE layer for a specific token $x$ can be described as below:
\begin{align}
    p(x) &= \frac{exp(g(x)_i)}{\sum^N_j exp(g(x)_j)}, \\
    y &= \sum_{i \in \mathcal{T}}p(x)E_i(x),
\end{align}
where $g(\cdot)$ refers to the sparse gating function, and $\mathcal{T}$ refers to the indices of top-$k$ values of $g(\cdot)$. The output of MoE is a linear combination of the computation of selected expert FFNs $f(\cdot)$. 

In expert parallelism, the parameters of experts do not share across workers, while those of other parts are identical across workers. Therefore, it is necessary to perform all-to-all communication across workers at the MoE layers in order to dispatch tokens to selected experts and combine them to their original experts. While \citet{gshard} and \citet{switch} implement the MoE on TPUs with one expert in each MoE layer on a TPU, we implement our model on Nvidia GPUs where there are several experts in each MoE layer on a GPU so as to fully utilize the memory. As all-to-all communication takes up a large amount of time, the optimization to improve efficiency is highly significant. We implement a series of optimization, including half-precision communication. A key problem is load balancing, which denotes that tokens can gather to only a few experts due to dynamic routing. Following \citet{switch}, we apply expert capacity, which refers to the number of tokens for an expert ($C = \frac{N \cdot c}{m}$, where $C$ refers to expert capacity, $N$ refers to the number of tokens in a batch, $c$ refers to capacity factor (which is a hyperparameter usually larger than $1.0$)  and $m$ refers to the number of experts), to alleviate this problem. Tokens out of the capacity of an expert are dropped from the computation and they are sent to next layers through residual connections. We find that the overloading problem can be severe, and this issue can be a significant one in the future research of expert models~\citep{base}. 

Besides the optimization in all-to-all communication, we compare the top-2 gating and top-1 gating and find that they can achieve similar model performance in perplexity, while the latter converges slightly slower. The effectiveness of top-1 gating enables faster computation. Besides, we also apply methods of memory optimization for higher efficiency. We find that gradient clipping globally can increase costs on all-to-all communication as it computes norms across all experts, and thus we apply local clipping for memory saving. We implement M6-100B with around 100 billion parameters on 128 Nvidia A100s and the speed of pretraining achieves 1440 samples/s (for samples of the sequence length of 272). 

We demonstrate that using MoE structure for model size scaling is effective and it can achieve similar performance to that of M6-10B, the largest dense model, within 2-3 times shorter time. The negative log perplexity of M6-100B reaches $-2.297$, in comparison with M6-10B that reaches $-2.253$ but with twice of time.\footnote{Note that the M6-10B trained on multimodal data has first been trained on plain text data, and it can actually start with much lower cross-entropy loss (around $1/3$ of the loss of the one trained from random initialization). We will make a more comprehensive comparison in order to fairly evaluate the effect and efficiency of the MoE scaling. } This shows that the MoE-based M6 model has advantages on the time basis compared with dense models with many more FLOPs.


\section{Applications} 
\subsection{Text-to-Image Generation}

\begin{figure*}
    \centering
    \includegraphics[width=\linewidth]{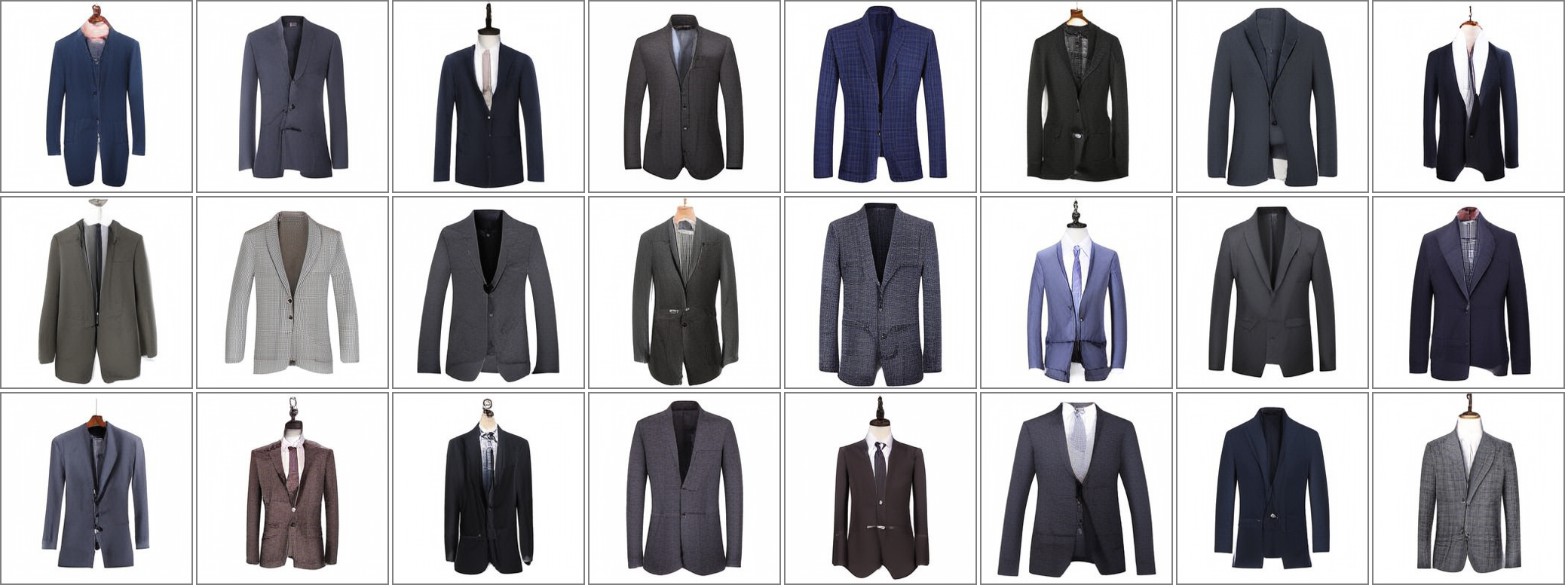}
    \caption{Generated images for \textit{sheep wool business casual suit} (\begin{CJK}{UTF8}{gbsn}绵羊毛商务休闲西服套装\end{CJK}).}
    \label{fig:suit}
\end{figure*}
\begin{figure*}
    \centering
    \includegraphics[width=\linewidth]{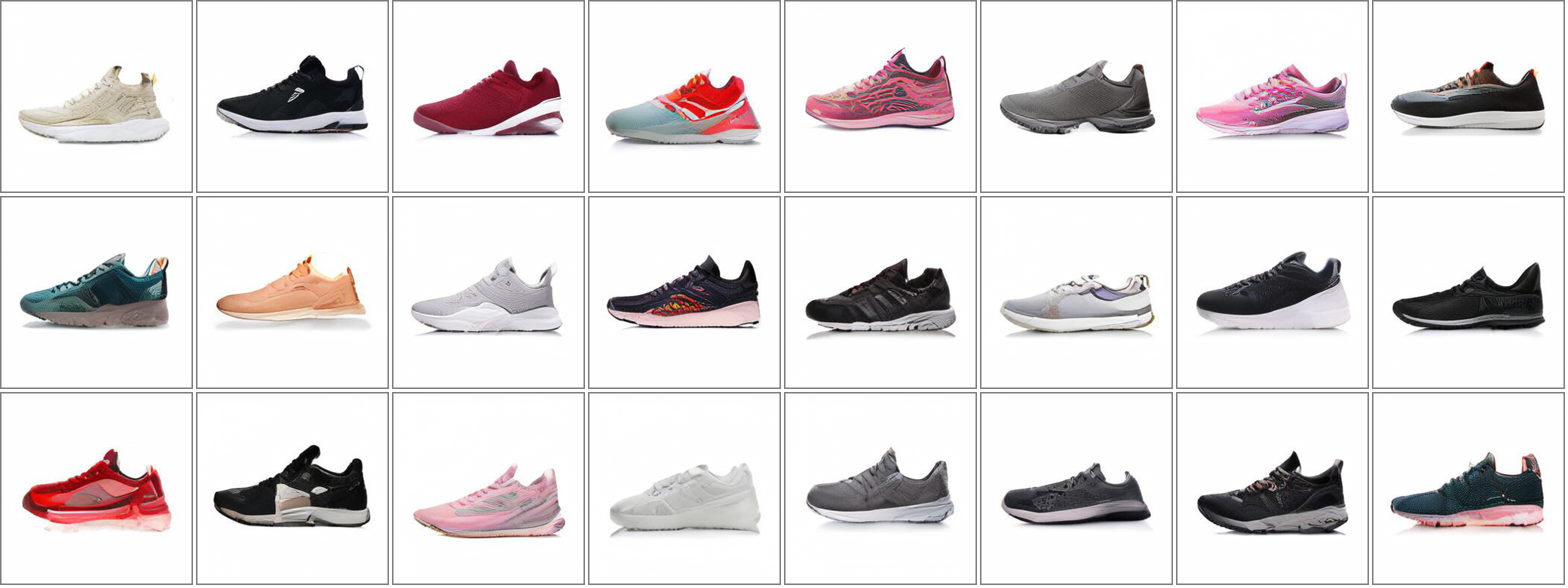}
    \caption{Generated images for \textit{shock absorption and breathable running shoes} (\begin{CJK}{UTF8}{gbsn}减震透气跑鞋\end{CJK}).}
    \label{fig:shoes}
\end{figure*}

\begin{figure*}
    \centering
    \includegraphics[width=\linewidth]{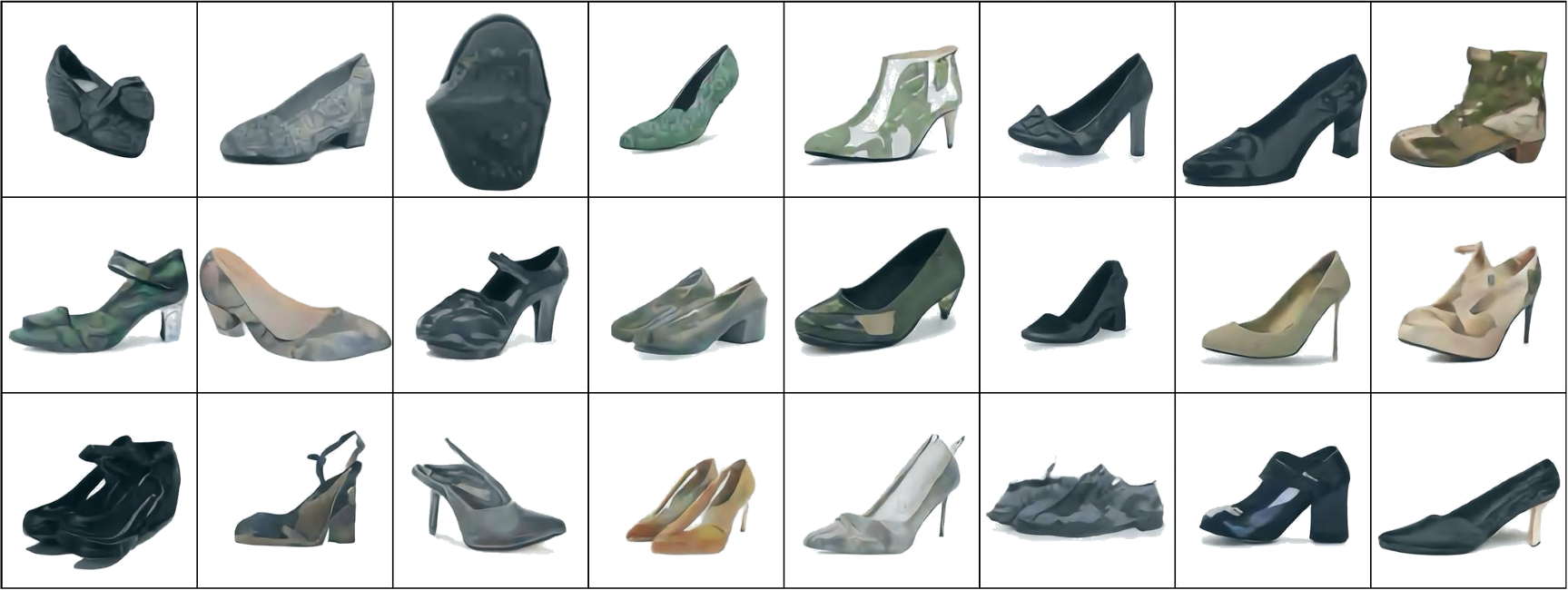}
    \caption{Generated images for \textit{military style camouflage high heels} (\begin{CJK}{UTF8}{gbsn}军旅风迷彩高跟鞋\end{CJK}).}
    \label{fig:highheels}
\end{figure*}
\begin{figure*}
    \centering
    \includegraphics[width=\linewidth]{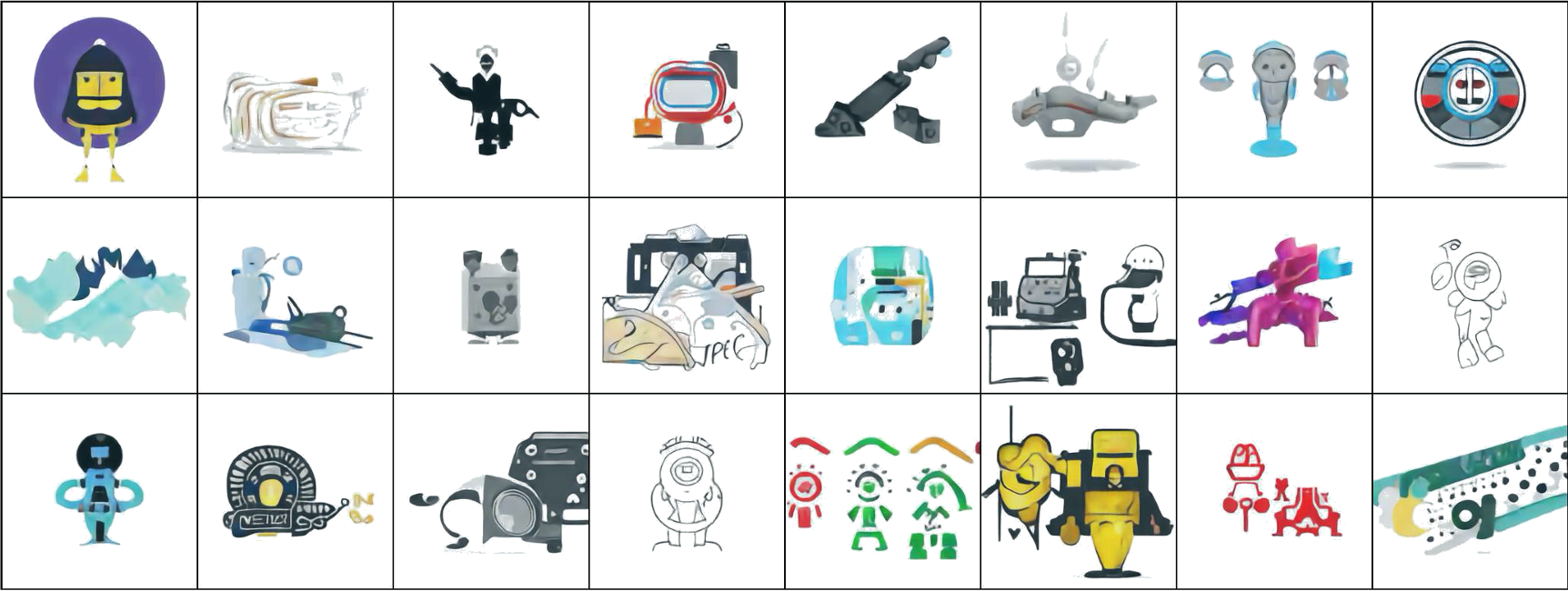}
    \caption{Generated images for \textit{a clip art of robots} (\begin{CJK}{UTF8}{gbsn}机器人矢量插图\end{CJK})).}
    \label{fig:robot}
\end{figure*}

Text-to-image generation has been an open problem for a long time. Previous studies mainly focused on generation on a limited domain, among which Generative Adversarial Nets (GANs)~\cite{xu2018attngan,goodfellow2014generative} are dominated methods.
Following~\citet{dalle}, we leverage a two-stage framework for text-to-image generation, including discrete representation learning and language modeling.

In the first stage, we focus on transforming images into sequences of discrete codes. 
There are a number of alternatives for discrete code generation, including VQVAE~\citep{vqvae} and VQGAN~\citep{vqgan}. 
In the second stage, it is necessary to build a language model to learn to generate text and code sequence. 
In the finetuning, we add code embedding and output layers to the pretrained M6. 
We concat the word sequence and the aforementioned generated code sequence as the input, and we set the objective of autoregressive language modeling for the training. 
At the stage of inference, we input the text sequence, and the model generates codes autoregressively with top-k sampling. 
The last step is to transform the code sequence to an image with the generator from the first stage. 

We construct a dataset for text-to-image generation in E-commerce.
Specifically, we collect over 50 million product titles and images from the mobile Taobao.
We apply a series of processing methods on the images to filter the unqualified.
We filter the images with complex background features (characters, patterns, etc.) with the in-house white-background image detector and OCR model.
We then filter the images with over 3 objects with our in-house object detector based on Faster R-CNN~\citep{faster-rcnn}. 
We finally obtain 1.8m high-quality product image-text pairs for finetuning. 
Compared with the images in the general domains, our collected data have the following features. 
The image and text are highly correlated as the text describes key features of the product, and there is no complex background in the images, which is easier to learn compared with the images in the public datasets such as MSCOCO~\citep{coco}.




We demonstrate two examples in Figure \ref{fig:suit} and Figure \ref{fig:shoes}. 
It can be found that the generated images have high quality and the generated objects resemble the real ones. 
Furthermore, in Figure \ref{fig:highheels} , we find that the model is able to imagine items according to the query \textit{military style camouflage high heels(\begin{CJK}{UTF8}{gbsn}军旅风迷彩高跟鞋\end{CJK})}, which do not exist in the real world.
The imagination ability provides room for creative design in real-world industrial scenarios, such as clothing design, shoe design, etc.


We also finetune M6 under our proposed framework on another dataset which contains 3 million images crawled from the Internet, which cover more general domains. 
And we find that the model can adapt to different domains. 
As shown in Figure~\ref{fig:robot}, the model is able to generate clip arts of robots .
This reveals the versatility of the framework in text-to-image generation. 

\subsection{Visual Question Answering}

We demonstrate our experimental results on a visual question answering dataset, and we illustrate how we directly apply the pretrained M6 to the VQA application. 

\begin{table}
  \caption{Results on the FMIQA dataset. We report both the overall accuracy and the accuracy on specific question types.} 
  \label{tab:fmiqa_result}
  \begin{tabular}{cccccc}
    \toprule
        Model   & Detection & Relation & Color & Number & Overall \\
        \midrule
        baseline    & 74.0  & 64.5  & 69.0  & 41.9  & 66.8  \\
        M6-base     & 79.0  & 71.0  & 70.9  & 45.2  & 71.0  \\
        M6-10B    & \textbf{83.0}  & \textbf{77.4}  & \textbf{72.7}  & \textbf{48.4}  & \textbf{74.7}  \\
    \bottomrule
  \end{tabular}
\end{table}

We leverage the FMIQA dataset \cite{fmiqa} as the Chinese visual QA benchmark, which requires the model to generate the answer given an image and a question. 
We implement a transformer-based model as our baseline. 
For the evaluation, we split the test set manually by random sampling 200 from the dataset as there is no official release of the test set, and we evaluate the overall accuracy by human evaluation. 
The results are demonstrated in Table~\ref{tab:fmiqa_result}. 
The pretrained M6-base outperforms the baseline by a large margin (+6.2\%), which indicates the effectiveness of multimodal pretraining. 
Scaling up the model to M6-10B further brings 5.2\% improvement. 

\begin{figure*}[tb]
    \centering
    \includegraphics[width=0.95\textwidth]{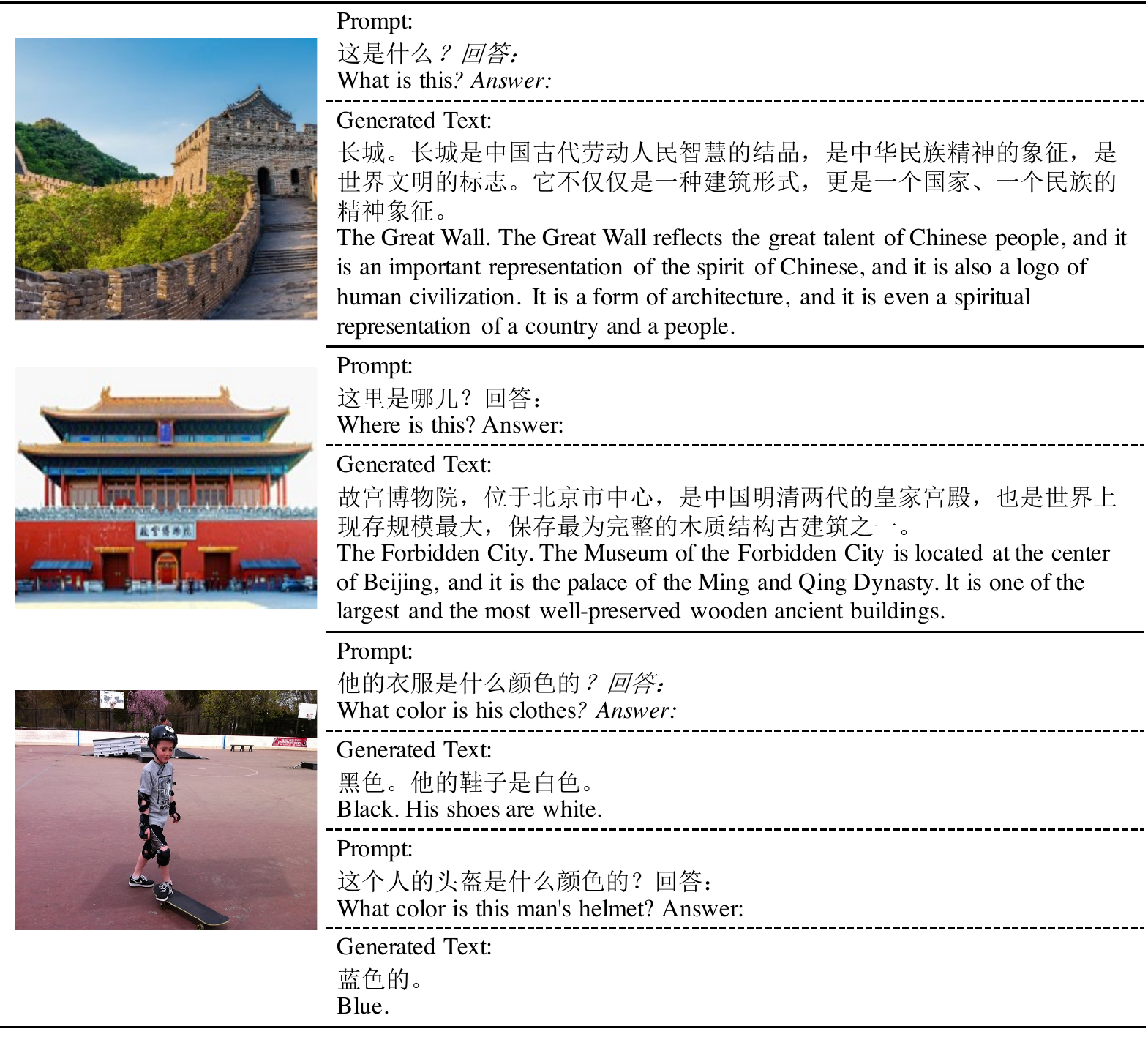}
    \caption{Several examples of general visual question answering without finetuning. We turn the origin questions to the designed pattern , with typical tokens such as ``?'' and ``Answer:''. 
    The pretrained model can recognize the question and provide the answer as well as some further description. 
    }
    \label{fig:i2t_case}
\end{figure*}

Furthermore, we show that simply finetuning on such a small VQA dataset may limit the potential of M6. Therefore, we directly leverage M6 for the VQA application. We find that the model is able to recognize general features and provide more related knowledge based on its understanding. Though the model pretrained on pseudo-parallel image-text pairs cannot directly answer questions about detailed features, such as color, number, etc., it is able to answer questions related to background knowledge. We demonstrate some examples in Figure~\ref{fig:i2t_case}.

\subsection{Image Captioning}

\begin{figure*}[tb]
    \centering
    \includegraphics[width=0.95\textwidth]{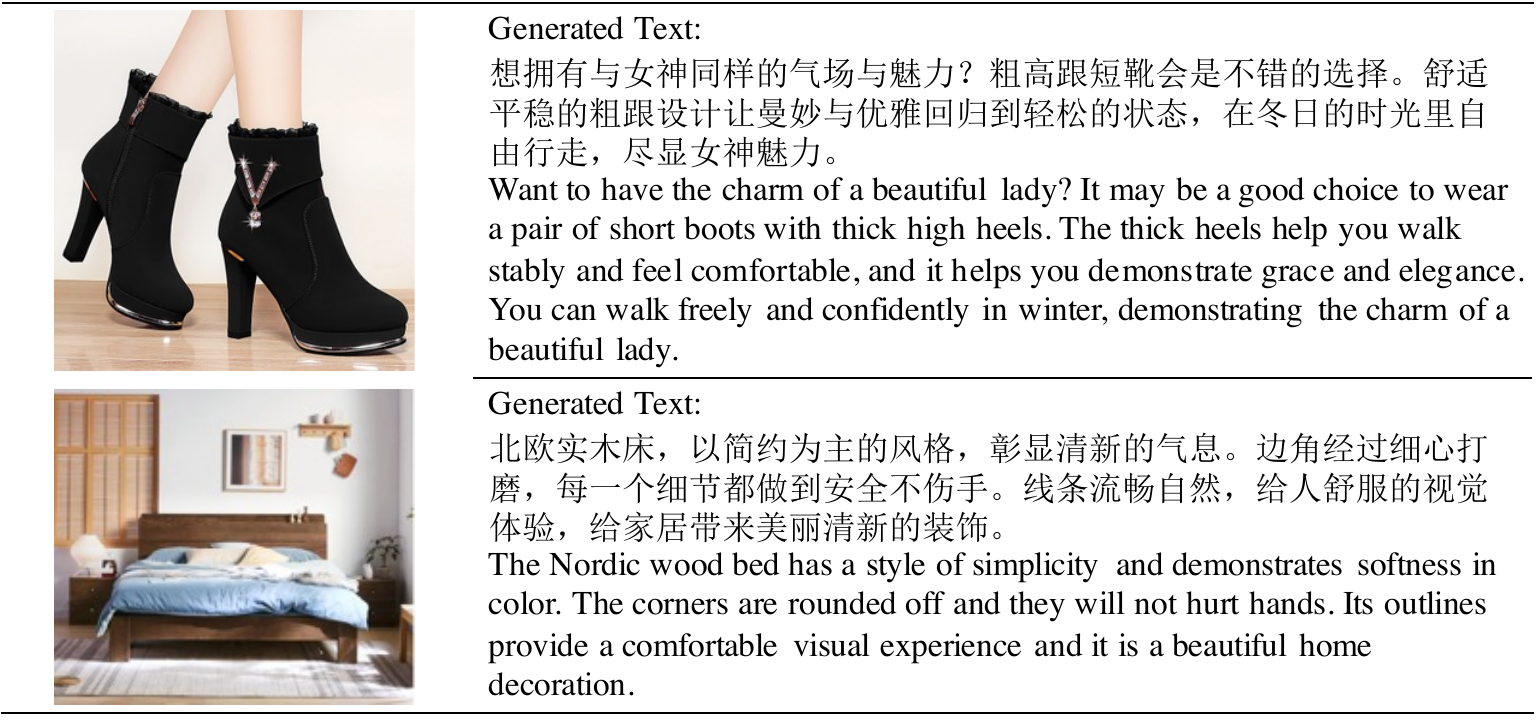}
    \caption{Two examples of image caption. We only use the image feature as input and provide no prompt. } 
    \label{fig:i2t_case3}
\end{figure*}

Image captioning requires the model to generate a caption that describes the given image, which examines the model ability of cross-modal generation. 
We construct a dataset (named E-Commerce IC) containing pairs of product descriptions and product images from Taobao. 
Since too long or too short descriptions may be noisy, we discard pairs with a description longer than 100 words or less than 10 words. 
To avoid dirty generations, we further use an in-house tool to filter descriptions that may contain dirty words (i.e., pornographic or violent words).
Finally, E-Commerce IC contains about 260k text-image pairs. 
We finetune the model with the image-to-text transfer task on E-Commerce IC.

We compare our model with a baseline of transformer in the human evaluation. 
We ask several annotators with the linguistic background to evaluate from three perspectives: grammar (whether a text is fluent without grammatical error), correctness (whether a text is faithful to the image), richness (whether a text is informative and attractive).
During the evaluation, we randomly sample 100 images from the test set.
For each image, an annotator is asked to score the text generated by different models.
The scores are within the range of $[0,5]$. 

The results in Table \ref{tab:ic_result} show that M6-base outperforms the baseline in all of the metrics. 
We find that all models achieve high scores in grammar.
However, in both correctness and richness, M6-base outperforms the baseline model by a large margin (+18.2\% and +14.4\%), indicating that multimodal pretraining helps to generate more faithful, informative and attractive texts. Scaling up the model to M6-10B further improves the correctness and richness (about 14.7\% and 7.0\%).
Figure \ref{fig:i2t_case3} illustrates two examples of image caption.
\begin{table}
  \caption{Results on the E-Commerce IC dataset.} 
  \label{tab:ic_result}
  \begin{tabular}{cccc}
    \toprule
        Model   & Grammar & Correctness & Richness \\
        \midrule
        baseline    & 4.45  & 2.58  & 3.12 \\
        M6-base    & 4.61  & 3.05  & 3.57 \\
        M6-10B   & \textbf{4.70} & \textbf{3.50} & \textbf{3.82} \\
    \bottomrule
  \end{tabular}
\end{table}

\subsection{Question Answering}

To demonstrate the potential availability in the applications of intelligent chatbots, we further employ the M6 model to generate long answers in the style of forum discussion.
Human-generated questions are collected from various Chinese forums, which are input to the model to generate the answer.
At the stage of inference, we append a question mark and a token \textit{``Answer:''} in the prompt, which better triggers the model to generate an answer.
To facilitate the generation of longer and more informative texts, we pick more complex questions.

\begin{figure*}[tb]
    \centering
    \includegraphics[width=0.95\textwidth]{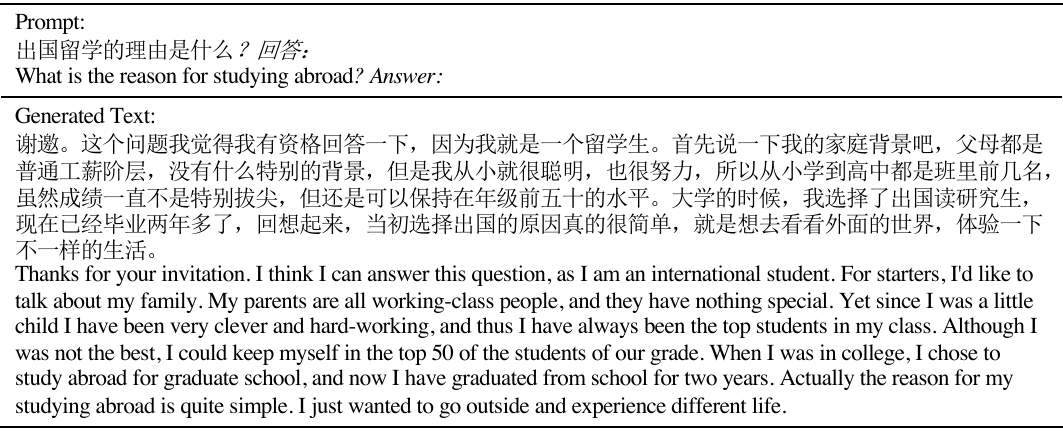}
    \caption{One example of general question answering. The prompt which includes the question successfully triggers the model to generate long texts in the style of forum discussion.}
    \label{fig:t2t_case}
\end{figure*}

Figure~\ref{fig:t2t_case} demonstrates an example of general question answering. 
The model can illustrate a man's own experiences that are related to the question and also point out the answer at the end. 
This generated text confused human annotators and passed the Turing Test. 
It shows that the model can not only  answer general questions  but also generate long fluency text. 

\subsection{Poem Generation}
We apply the pretrained model to Chinese poem generation. 
The model is able to generate genres with format constraints.

\begin{figure}
    \centering
    \includegraphics[width=\linewidth]{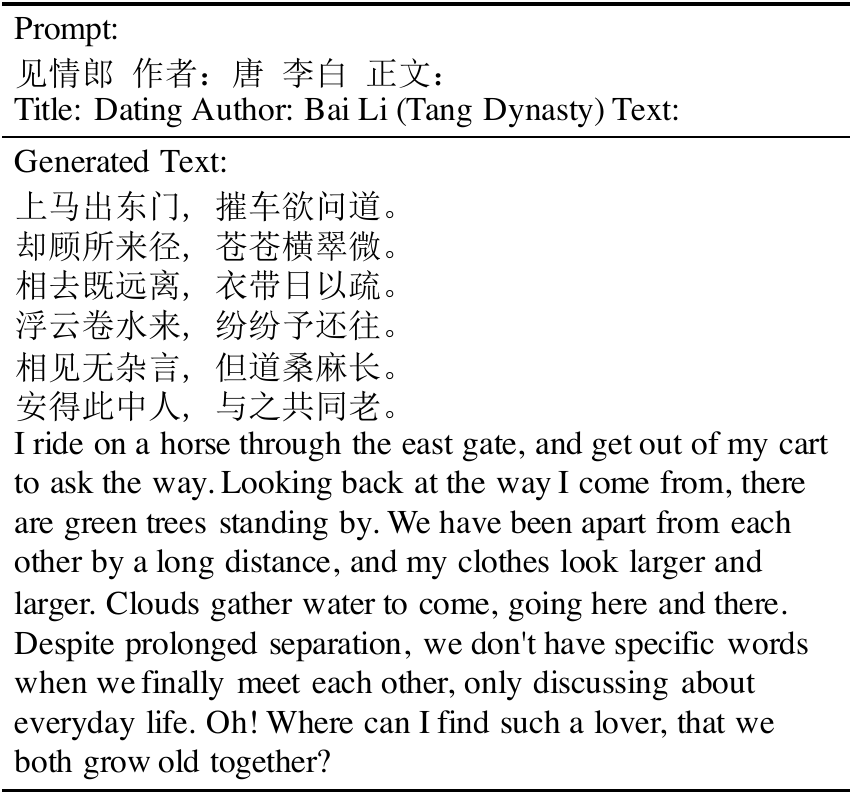}
    \caption{One example of a generated poem, the prompt and the constraint mask work together to generate a poem based on the given title.}
    \label{fig:poem}
\end{figure}
Ancient Chinese poetry has various specific formats. We adopt the simplest constraints that
\begin{itemize}
    \item The poem shall be consisted of at least 4 lines.
    \item The total number of lines shall be even.  
    \item Each line must have exactly 5 or 7 words.
    \item All lines shall have the same number of words.
\end{itemize}


Text generation under format constraint is done in a search framework that we generate short sentences ending with punctuation until the number of words meets the constraint.
We repeat this process until the model generates an "\textit{<eos>}" token, or the number of lines exceeds a limit of 16.
Figure \ref{fig:poem} illustrates an example of a generated poem.

    
        
        









\subsection{Image-Text Matching}

We evaluate the model's ability in cross-modal retrieval. 
Specifically, we construct a dataset (named E-Commerce ITM) containing pairs of texts and images from the mobile Taobao. 
Each pair belongs to a single item. 
we collect 235K products in the clothing industry from Taobao.
For each product, aside from the product image, we obtain a query by rewriting the product title. 
Specifically, we conduct named entity recognition on the title using an in-house tool, which extracts the terms describing the style, color, category and texture of the product. These terms are 
then concatenated into a natural language query, which is used in image-text matching. The length of each query is between 6 to 12 words. 
The pairs of the query and corresponding product image are labeled as positive samples. The negative samples are constructed by randomly substituting the query in the original pairs.

We require the model to perform binary classification to discriminate positive and negative samples. 
We compare our model with InterBert~\cite{interbert}, which is also a Chinese multi-modal pretrained model effective in cross-modal classification downstream tasks. The InterBert utilizes object-based features and has been pretrained on Taobao product image-text data as well.

The results are shown in Table~\ref{tab:itm_result}. 
It should be noted that the InterBert and M6-base are both implemented with transformer-based architecture and have similar model scales. 
However, M6-base still outperforms InterBert by 10.3\%.
In experiments, we find the product images generally contain relatively fewer detected objects, which may harm the performance on this task. 
In contrast, M6 avoids this problem by employing the patch features and achieves much better performance.

\begin{table}
  \caption{Results on the E-Commerce ITM dataset. We report the accuracy on the test set.} 
  \label{tab:itm_result}
  \begin{tabular}{ccc}
    \toprule
        Model   & Accuracy & Improvement \\
        \midrule
        InterBert & 81.8  & - \\
        M6-base     & \textbf{90.2}  & 10.3\% \\
    \bottomrule
  \end{tabular}
\end{table} 

\section{Related work}
\label{sec:related-work}

The tremendous success of NLP pretraining, including BERT~\citep{bert}, GPT~\citep{gpt, gpt2, gpt3}, and also some other related studies~\citep{xlnet, albert, roberta, unilm, unilmv2}, inspires the research in cross-modal representation learning. Also, recent studies show that the ubiquitous Transformer architecture~\citep{transformer} can be extended to different fields, including computer vision~\citep{vit, detr}. Therefore, the simplest solution to incorporate recent pretraining methods and cross-modal representation learning is the extension of BERT. From the perspective of architecture, there are mainly two types, including single-stream model and dual stream model. Specifically, single-stream model is simple and it gradually becomes the mainstream architecture. These models mostly differ in their designs of pretraining tasks or the construction of input image features. 
Basically, they are mainly pretrained masked language modeling, masked object classification, and image-text matching. VisualBERT~\citep{visualbert} and Unicoder-VL~\citep{unicoder-vl} simply use BERT and are pretrained with the aforementioned tasks. UNITER~\citep{uniter} pretrains the model with an additional task of word-region alignment. Oscar~\citep{oscar} enhances the alignment between objects and their corresponding words or phrases. VILLA~\citep{villa} further improves model performance by adding their proposed adversarial learning methods to pretraining and finetuning. Except for pretraining tasks, some studies focus on the features of images. Most pretraining methods for multimodal representation learning utilize the features generated by a trained object detector, say Faster R-CNN~\citep{faster-rcnn}. PixelBERT~\citep{pixelbert} accepts raw images as input and extract their latent representations with a learnable ResNet~\citep{resnet} or ResNext~\citep{resnext}. FashionBERT~\citep{fashionbert} splits the images into patches with a trained ResNet without co-training. Besides single-stream models, dual-stream models also can achieve outstanding performance, such as VilBERT~\citep{vilbert}, LXMERT~\citep{lxmert} and InterBERT~\citep{interbert}. ViLBERT-MT~\citep{vilbert-mt} enhances model performance with multi-task finetuning. ERNIE-ViL~\citep{ernie_vil} enhances the model with the application of scene graph information. 
In spite of these successful cases, it still requires further researches to unmask the success of multimodal pretraining.

\section{Conclusions}

In this work, we propose the largest dataset M6-Corpus for pretraining in Chinese, which consists of over 1.9TB images and 292GB texts. 
The dataset has large coverage over domains, including encyclopedia, question answering, forum discussion, common crawl, etc. 
We propose a method called M6 that is able to process information of multiple modalities and perform both single-modal and cross-modal understanding and generation.  
The model is scaled to large model with 10B and 100B parameters with sophisticated deployment, and both models are the largest multimodal pretrained models. 
We apply the model to a series of downstream applications, showing its versatility. 
More specifically, we design a downstream task of text-guided image generation, and the finetuned M6 can reach superior performance by producing images of high quality. 

In the future, we will continue the pretraining of extremely large models by increasing the scale of data and models to explore the limit of performance, and we also endeavor to search for more downstream applications for further generalization.



\balance
\bibliographystyle{ACM-Reference-Format}
\bibliography{reference}


\clearpage 


\end{document}